\PassOptionsToPackage{unicode}{hyperref}
\PassOptionsToPackage{hyphens}{url}
\documentclass[
]{article}
\usepackage{amsmath,amssymb}
\usepackage{iftex}
\ifPDFTeX
  \usepackage[T1]{fontenc}
  \usepackage[utf8]{inputenc}
  \usepackage{textcomp} 
\else 
  \usepackage{unicode-math} 
  \defaultfontfeatures{Scale=MatchLowercase}
  \defaultfontfeatures[\rmfamily]{Ligatures=TeX,Scale=1}
\fi
\usepackage{lmodern}
\ifPDFTeX\else
\fi
\IfFileExists{upquote.sty}{\usepackage{upquote}}{}
\IfFileExists{microtype.sty}{
  \usepackage[]{microtype}
  \UseMicrotypeSet[protrusion]{basicmath} 
}{}
\makeatletter
\@ifundefined{KOMAClassName}{
  \IfFileExists{parskip.sty}{%
    \usepackage{parskip}
  }{
    \setlength{\parindent}{0pt}
    \setlength{\parskip}{6pt plus 2pt minus 1pt}}
}{
  \KOMAoptions{parskip=half}}
\makeatother
\usepackage{xcolor}
\NewDocumentCommand\citeproctext{}{}
\NewDocumentCommand\citeproc{mm}{%
  \begingroup\def\citeproctext{#2}\cite{#1}\endgroup}
\makeatletter
 \let\@cite@ofmt\@firstofone
 \def\@biblabel#1{}
 \def\@cite#1#2{{#1\if@tempswa , #2\fi}}
\makeatother
\newlength{\cslhangindent}
\newlength{\csllabelwidth}
\newenvironment{CSLReferences}[2] 
 {\begin{list}{}{%
  \setlength{\itemindent}{0pt}
  \setlength{\leftmargin}{0pt}
  \setlength{\parsep}{0pt}
  \ifodd #1
   \setlength{\leftmargin}{\cslhangindent}
   \setlength{\itemindent}{-1\cslhangindent}
  \fi
  \setlength{\itemsep}{#2\baselineskip}}}
 {\end{list}}
\usepackage{calc}

\ifLuaTeX
  \usepackage{selnolig}  
\fi
\usepackage{bookmark}
\IfFileExists{xurl.sty}{\usepackage{xurl}}{} 
\hypersetup{
  pdftitle={Deep learning for echo sounder data},
  pdfauthor={Ketil Malde},
  hidelinks,
  pdfcreator={LaTeX via pandoc}}

\title{Deep learning for echo sounder data}
\author{Ketil Malde}
\date{2026}

\begin{document}
\maketitle

\subsection{\texorpdfstring{\emph{Abstract}}{Abstract}}\label{abstract}

\emph{There is no doubt that over the last decade, techniques from the
field of machine learning have revolutionized how we process and
interpret data, especially images and text. For underwater observations
acoustics is a primary source of information, and naturally, deep
learning methods have been applied to echograms and other acoustics
data, but so far with rather modest results. Here, we argue that due to
intrinsic properties of acoustic data, substantial advances will likely
require research into deep learning methods beyond mere recycling of
models and techniques from image processing. Currently, the potential
for breakthroughs in method development is hindered by the lack of
standard data formats and organization, and even more by the lack of
readily available, high quality data sets with established performance
goals. To advance the field, these shortcomings should be remedied.}

\subsection{Introduction}\label{introduction}

Acoustics give scientists a unique way to observe the ocean beneath the
surface. In particular, split beam echo sounders have become a standard
tool in marine science, useful both to quantify biomass in a volume of
water and able to detect, locate, and distinguish individual fish or
other objects in the water column. The data produced can be visualized
in the form of echograms which traditionally are scrutinized by experts
to extract ecological and oceanographic information.

Machine learning, and in particular deep learning
(\citeproc{ref-goodfellow2016deep}{Goodfellow, Bengio, and Courville
2016}), is a set of powerful techniques that can automate this kind of
scrutiny. Consequently, a number of deep learning models have been
proposed to perform acoustic target classification, that is, interpret
echograms and classify their contents.

\subsection{Deep learning for acoustic target
classification}\label{deep-learning-for-acoustic-target-classification}

Traditional machine learning methods have been used to support target
classification for a long time. One example is the SHAPES
(\citeproc{ref-coetzee2000use}{Coetzee 2000}) method offered by Echoview
(\citeproc{ref-EchoviewSoftware}{Echoview Software Pty Ltd 2025}),
another is the feature library used by the KORONA analysis package
(\citeproc{ref-korneliussen2016acoustic}{Korneliussen et al. 2016}).
With the advent of deep learning, and especially in the wake of its
remarkable success in image processing, several recent works have
applied deep learning methods to echograms. A common approach is to
split echograms into tiles and use a deep neural network classifier to
categorize them (\citeproc{ref-hirama2017discriminating}{Hirama et al.
2017}; \citeproc{ref-rezvanifar2019deep}{Rezvanifar et al. 2019}).
Recently, Pala et al. (\citeproc{ref-pala2024self}{2024}) used tiled
data to use self-supervised training. Other works have applied U-Net
(\citeproc{ref-ronneberger2015u}{Ronneberger, Fischer, and Brox 2015})
architectures to process larger echogram regions and produce fine
grained segmentation masks that define regions that represents fish
(\citeproc{ref-brautaset2020acoustic}{Brautaset et al. 2020}) or krill
(\citeproc{ref-li2025enhancing}{Li et al. 2025};
\citeproc{ref-bahlburg2025mapping}{Bahlburg et al. 2025}). Object
detection models, including YOLO v2
(\citeproc{ref-redmon2017yolo9000}{Redmon and Farhadi 2017}) and Faster
R-CNN (\citeproc{ref-ren2016faster}{Ren et al. 2016}) have been used to
locate and classify schools (\citeproc{ref-marques2020schools}{Marques
et al. 2020}). See also the review by
(\citeproc{ref-yassir2023acoustic}{Yassir et al. 2023}).

\subsection{Discussion}\label{discussion}

Based on our cursory sample of the literature, we see that applications
of deep learning so far tend to use image-based methods based on
convolutional neural networks (CNNs) for applications which target
narrow data sets and specific classification problems. This raises the
question of to what extent methods designed for image analysis are the
best tool for acoustic data processing, and if not, how we best can
stimulate new innovation to advance the field.

\subsubsection{Are convolutional networks the right
tool?}\label{are-convolutional-networks-the-right-tool}

CNNs have over the last fifteen years been enormously successful for
image analysis. Like all machine learning methods, CNNs have an
\emph{inductive bias}, that is, they make implicit assumptions about the
data that allow them to generalize from the specific (i.e.~training
data) to the general (i.e.~new samples to be classified).

Specifically, the convolutions that make up a CNN are
\emph{translationally equivariant.} That is, detections depend on local
structures of the input, but not on its location. For images, this is a
desirable property, an object looks more or less the same regardless of
placement, and can usually be expected to occur with the same
probability anywhere in the picture. Another benefit is that by
constraining the inputs to a small region, the number of parameters of
the system can be drastically reduced.

An echogram is not an image, however. For acoustics, the assumption that
features occur with the same probability everywhere does not hold.
Certain targets occur in specific places, for instance, there will
usually be one bottom (or none), one surface, and fish and other marine
life in between. The various species tend to occur at specific depth
ranges, which can vary with the time of day and with season, and so on.

Moreover, the acoustic signal itself will change its characteristics
with increasing range (which is not the same as depth). As range
increases, the cone widens, and the observed volume grows. At short
ranges, individual fish can be resolved as sharp points, at longer
ranges, the echo from each fish will be smeared over a larger area, and
with sufficient fish densities, the echo from multiple fish will be
intermixed. And while reflected light captured by a camera travels
relatively unhindered through the air, the emitted sound from a
transducer will attenuate as it travels through the water column, and
the attenuation will vary depending on physical variables and the
presence or absence of biological and other scattering matter.

\subsubsection{Is segmentation and object detection what we
want?}\label{is-segmentation-and-object-detection-what-we-want}

For images, it is often of interest to break the contents down into a
set of objects. Segmentation assigns a class to contiguous but often
irregular regions of the image, while object detection usually produces
bounding boxes around objects of interest. In acoustics, echograms have
traditionally been annotated by dividing them into non-overlapping
regions (layers or schools) that are assigned species labels, and it is
tempting to apply segmentation algorithms (like U-Net) or object
detection (like YOLO or R-CNN) to echograms as well.

While an echogram can often usefully be broken down into regions, this
does not adequately capture the underlying truth. Species often occur in
densities that vary continuously across space, and while specific
species tend to occur in layers, these layers can shift and intermix.
Moreover, for many image segmentation applications, it is important to
precisely find object boundaries, and specific loss functions have been
developed with this in mind
(\citeproc{ref-bokhovkin2019boundary}{Bokhovkin and Burnaev 2019}). For
acoustics, region boundaries can be vague or gradual, and annotated
boundaries can be set liberally as long as they enclose a fish school or
other region of interest.

\subsubsection{What are appropriate augmentation
methods?}\label{what-are-appropriate-augmentation-methods}

In order to increase the diversity of the training data, it is common to
apply transformations that are assumed to preserve class (or other
information we want to extract). This process is called data
\emph{augmentation}. For images, we can for instance modify the colors,
to simulate how a scene would look under different light conditions.
Another common augmentation technique is to rotate the image (usually by
a small angle), crop it, or mirror it.

Common machine learning frameworks provide easily accessible
augmentation methods, but care must be taken that the specific
transforms are appropriate for acoustic data. Underwater structures that
are subject to translation or rotation may not realistically represent
the same class. It can be tempting to treat multiple frequencies as
color channels, but where a color transform can make sense for objects
in a photo, for acoustic classification, the precise relationship
between the backscatter at different frequencies are in fact a
distinguishing feature. Generally speaking, many image augmentation
methods assume that objects are identified from their shape and should
be robust to changes to other characteristics, while this is less often
true for acoustic data.

Conversely, acoustic data come with its own specific errors and
characteristics, which effective augmentation methods need to take into
account. Such augmentation methods would include simulating noise and
common artifacts to capture variability present in real data, but
without changing the categories or other information we want to extract.

\subsubsection{Can we obtain sufficient annotations for supervised
training?}\label{can-we-obtain-sufficient-annotations-for-supervised-training}

The most successful methods are \emph{supervised}, that is, they are
trained from data samples paired with the desired output. For images,
this means labels assigned by an annotator, or labels with a location or
object mask for object detection and segmentation. For text, this is
usually in the form of predicting the next word from the previous ones,
or by having paired questions and answers, or translations.

For acoustics data, a reliable ground truth can be difficult to obtain,
and most annotations are based on the subjective judgement by experts
with extensive experience. The use of multiple frequencies and phase can
provide useful information but increases the complexity of the data.
Current broadband echo sounders capture large amounts of information in
the form of acoustic spectra which are not readily visualized as an
echogram image, and difficult to interpret even for experts. At the same
time, it is difficult to obtain ancillary information that can provide
labels. Trawl sampling suffer from selectivity as small organisms pass
through and large specimens avoid the trawl entirely, and in any case
can only provide aggregated data over the haul, and not high spatial
resolution. Trawl cameras can mitigate these issues somewhat, but as it
is towed some distance behind the ship, observations are delayed and
fish will have time to react in ways that make the camera observations
different from what of the echo sounder.

\subsection{Needs on the way forward}\label{needs-on-the-way-forward}

CNNs and general deep learning technologies developed for images are
tremendously powerful, and although acoustics is a different kind of
data, the performance can quite possibly be sufficient for many
purposes. Yet, there is reason to believe that developing specialized
methods that are tailored to the characteristics of the data could
improve performance on specific tasks and generalization.

\subsubsection{Specialized deep learning models, methods, and
architectures}\label{specialized-deep-learning-models-methods-and-architectures}

Over the last decade, research into deep learning has given us a diverse
set of powerful tools and techniques. The two overwhelmingly dominant
data types that have driven innovation are images and text, and one
could perhaps say the technological evolution is overfitting to these
data types. If we wish to see similar gains for other data types, we
need research into new model architectures, augmentation techniques,
output formats, and possibly cost functions and training regimes.

A reader who is familiar with the workings of CNNs will probably object
at this point. Variables like depth and range, temperature, time of day,
latitude, and so on, can easily be encoded into the input of a CNN, for
instance as a separate input layer. And if the inherent properties of a
CNN architecture fails to capture sufficient context, there exist other
architectures that do. For instance can recurrent networks (RNNs)
``remember'' context, and e.g.~a bidirectional RNNs can ``know'' how far
it is to the surface or sea bed, or any other feature it has detected in
the water column. Similarly, transformer architectures (which have
largely replaced RNNs for many applications) incorporate attention that
allows the model to combine information from different locations in the
data. A similar argument can be made for augmentation techniques, where
there exists options like noise injection, horizontal flips, and
cropping which may be more appropriate.

Although the limitations of image-oriented methods can be mitigated
somewhat, there exists uncharted opportunities for exploration and
likely new discoveries. Similar to how vision transformers
(\citeproc{ref-dosovitskiy2020image}{Dosovitskiy et al. 2020}) were an
innovative adaptation of an architecture originally developed for
sequential data, the potential for similar unforeseen adaptations or
innovations exists for acoustic data. Their invention will require both
deep understanding of the problem domain, the characteristics of the
data, and machine learning theory.

\subsubsection{Well organized training data available to the
public}\label{well-organized-training-data-available-to-the-public}

The importance of standardized data sets like ImageNet
(\citeproc{ref-deng2009imagenet}{J. Deng et al. 2009}), MNIST
(\citeproc{ref-deng2012mnist}{L. Deng 2012}), Pascal VOC
(\citeproc{ref-everingham2010pascal}{Everingham et al. 2010}), and COCO
(\citeproc{ref-lin2014microsoft}{Lin et al. 2014}) can hardly be
overstated. These data sets serve to define what constitutes important
challenges in the field, supply training data, and provide universally
agreed upon benchmarks. The real breakthrough for image classification
came in 2012, when Krizhevsky, Sutskever, and Hinton
(\citeproc{ref-krizhevsky2012imagenet}{2012}) demonstrated not only the
efficacy of convolutional neural networks, but also of techniques like
dropout, augmentation, ReLU activation, max pooling, and GPU training.
Although these innovation have earlier precursors
(\citeproc{ref-schmidhuber2022scientific}{Schmidhuber 2022}), it was the
unprecedented performance on the ImageNet data set that established CNNs
as an undisputed paradigm shift. The value of later algorithmic advances
have been similarly been established by improved benchmark scores, for
instance residual connections (\citeproc{ref-he2016deep}{He et al.
2016}) and visual transformers
(\citeproc{ref-dosovitskiy2020image}{Dosovitskiy et al. 2020}) for
ImageNet, and single-pass object detection
(\citeproc{ref-redmon2016you}{Redmon et al. 2016}) and region proposals
(\citeproc{ref-ren2016faster}{Ren et al. 2016}) for VOC and COCO.

In contrast, the lack of robust, large, and standardized data sets to
train and evaluate models is the primary obstacle to developing deep
learning methods specifically for acoustics. Such data sets need to
represent important analysis tasks, capture the full variability
inherent in the data distribution and, for supervised learning, come
with an established ground truth and performance measures.

Data diversity is needed as deep learning methods are often fragile, and
small changes in the input data can cause large reductions in accuracy.
Adversarial examples (\citeproc{ref-szegedy2013intriguing}{Szegedy et
al. 2013}) can often be constructed, where imperceptible (to a human)
changes lead to severe misclassification, but also natural changes in
the data domain can cause reliability to decay. For results to be
generalizable, training and benchmarking data sets need to capture as
much of the inherent variability as possible. This means sampling over
extensive time periods, geographical areas, time and season, and using a
variety of equipment and configurations.

Equally important is the existence of robust and agreed upon ground
truth annotations that accurately represent the information we want to
extract. In practice, annotations are often made by subjective judgement
of experts, in combination with auxiliary information like trawl
catches. But if objectively correct ground truth annotations can be
obtained, it is possible for automated methods to exceed the performance
of experts.

\subsubsection{Standardized processing and data
formats}\label{standardized-processing-and-data-formats}

A final obstacle is the large variety in equipment, data formats, and
processing software. Modern echo sounders like the Kongsberg EK80 offer
many configuration choices, and a data file can contain data where the
configuration varies between individual pings. The rapid development in
equipment features is reflected in subtle changes in the data formats,
and unless care is taken, can result in subtle incompatibilities. To
convert raw data into formats that are more easily processed (typically
regular grids with backscatter converted to a standard unit like Sv),
software packages (e.g.~Echopype, pyEchoLab, Korona/LSSS) are available,
but these make implicit and different choices for regridding,
interpolation, and other processing steps, and their output file formats
(e.g.~for annotations) differ. There are also choices of coordinate
systems to use. Is the y-axis depth or range? Is it corrected for the
motion of the ship? Is the x-axis ping number or a time stamp, and if
so, using what representation? Each option has advantages and
disadvantages, but the specific choice is less important than
consistency, and ensuring that the acoustic data, annotations, and any
other information are robustly interoperable.

Some of this complexity can be unavoidable, and there are different use
cases that require echo sounder data in the form of Sv, TS,
spectrograms, pulse compressed, or just the raw captured signal.
Nevertheless, any additional, unnecessary complexity adds needlessly to
the confusion for any machine learning expert that attempts to grapple
with it. There have been efforts to improve interoperability by
standardizing data formats, like HAC
(\citeproc{ref-mcquinn2022description}{McQuinn and Reid 2022}) and the
Sonar-netCDF4 standard (\citeproc{ref-macaulay2018sonar}{Macaulay and
Peña 2018}), and more recently by defining processing flows and stages
(``echo levels''). However, such initiatives have not so far lead to
wide adoption. At this point, it is probably better to publish data in
formats that are convenient for the publisher and existing users (which
are often the same person or institution). Reuse and interoperability
can be served by documenting the choices made, and by providing working
code for accessing and manipulating the data.

\subsection{Conclusion}\label{conclusion}

The importance of acoustic data in marine science, together with the
large volumes of data produced, make acoustics an exciting frontier for
new machine learning research. For more than a decade, deep learning has
produced a series of tremendous successes in image and text analysis.
Attempts to transfer these technologies to marine acoustics have
resulted in comparatively modest gains, and the potential for truly
innovative developments remains.

To fully unleash the power of machine learning requires solid domain
knowledge built from years of experience in combination with
foundational competence in machine learning theory and practice. It is
not enough simply to hand an acoustics data set to a computer scientist,
without a deeper understanding of the data she will treat it like the
image problem it resembles. Similarly, an acoustic scientist scratching
the surface of machine learning will discover the very successful image
classification models and apply them without considering their
limitations or investigating other options. A true cross disciplinary
research environment is needed, but takes long term commitment and
substantial investment to build.

The lack of established training data and benchmarks remains a major
obstacle. Without setting clear and quantitative goals that are
representative of the field in general, it becomes hard or even
impossible to measure the true gains. To motivate as well as to monitor
progress, the acoustics community needs to construct its own versions of
ImageNet. To stimulate and accelerate innovation, the community must
first define clear challenges with representative and quantitative
performance measures, second, provide adequate data sets in sufficiently
portable and standard formats and ensure their wide availability, and
finally, develop the cross disciplinary research environments with the
necessary competence and experience, both in machine learning and in
marine acoustics. Once these components are in place, the field will be
open to new and innovative approaches that in turn will be instrumental
to better understand our oceans.

\subsection*{References}\label{references}
\addcontentsline{toc}{subsection}{References}

\phantomsection\label{refs}

\end{document}